\documentclass[11pt,a4paper]{article}
\usepackage[hyperref]{ranlp2023}

\usepackage{graphicx}
\usepackage{multirow}
\usepackage{footnote}
\usepackage{ragged2e}
\makesavenoteenv{tabular}
\makesavenoteenv{table}
\usepackage{hyperref}
\usepackage{tabularx,booktabs, longtable}
\usepackage{caption,subcaption}
\usepackage{svg}
\usepackage{color}
\usepackage{tcolorbox}
\usepackage{CJK}
\usepackage{adjustbox}
\usepackage{array}    
\usepackage{times}
\usepackage{latexsym}
\usepackage{longtable}
\usepackage{amsfonts}
\usepackage{amsmath}

\usepackage{microtype}
\usepackage[utf8]{inputenc}

\aclfinalcopy 

\title{BanglaBait: Semi-Supervised Adversarial Approach for Clickbait Detection on Bangla Clickbait Dataset}

\author{Md. Motahar Mahtab \\
  BRAC University\\ 
  Dhaka, Bangladesh \\
  \texttt{mahtab27672767@gmail.com} \\
  \And
  Monirul Haque \\
  BRAC University\\ 
  Dhaka, Bangladesh \\
  \texttt{monirul.haque.mail@gmail.com}
  \AND
  Mehedi Hasan \\
  BRAC University\\ 
  Dhaka, Bangladesh \\
  \texttt{mehedi.hasan@g.bracu.ac.bd} \\
  \And
  Farig Sadeque \\
  BRAC University\\ 
  Dhaka, Bangladesh \\
  \texttt{farig.sadeque@bracu.ac.bd} }

\date{}

\begin{document}
\maketitle
\begin{abstract}
Intentionally luring readers to click on a particular content by exploiting their curiosity defines a title as clickbait. Although several studies focused on detecting clickbait titles in English articles, low-resource language like Bangla has not been given adequate attention. To tackle clickbait titles in Bangla, we have constructed the first Bangla clickbait detection dataset containing 15,056 labeled news articles and 65,406 unlabelled news articles extracted from clickbait-dense news sites. Each article has been labeled by three expert linguists and includes an article's title, body, and other metadata. By incorporating labeled and unlabelled data, we finetune a pre-trained Bangla transformer model in an adversarial fashion using Semi-Supervised Generative Adversarial Networks (SS-GANs). The proposed model acts as a good baseline for this dataset, outperforming traditional neural network models (LSTM, GRU, CNN) and linguistic feature-based models. We expect that this dataset and the detailed analysis and comparison of these clickbait detection models will provide a fundamental basis for future research into detecting clickbait titles in Bengali articles. We have released the corresponding code and dataset \footnote{\url{https://github.com/mdmotaharmahtab/BanglaBait}}.
\end{abstract}

\section{Introduction}\label{sec:introduction}
Due to the widespread usage of the internet, the news industry has progressively evolved into an online news industry leading to the explosion of clickbait titles in recent years. As the concept of clickbait can be hazy to grasp, the classification of clickbait is a highly subjective endeavor. \citet{biyani20168} suggests that clickbait titles can be roughly categorized into eight types. Table \ref{table:Clickbait categories} displays this different clickbait categories\footnote {'wrong' category in \citet{biyani20168} was replaced by 'question' category - reason described in details in section \ref{sec:data_collection}} and their corresponding Bangla articles.

\begin{table*}[!hbt]
\centering
\small
\begin{tabular}{lp{0.18\linewidth}p{0.6\linewidth}} 
  \hline
  \textbf{Category} & \textbf{Reason} & \textbf{Headline example \& Translation}  \\
  \hline

  Questions & Titles pose a query that compels the reader to click to get the answer. &  \raisebox{-.26\height}{\includegraphics{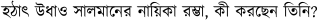}} (Salman's actress Rombha mysteriously disappeared, what is she up to?). \\

  Inflammatory & Titles evoke strong emotion. &  \raisebox{-.26\height}{\includegraphics{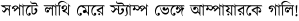}} (Lost his cool, kicked, then busted out the stamps before abusing the umpire!) \\

  Curiosity Gap/Teasing & Titles leave the reader in the dark, which tempts them to click. &  \raisebox{-.26\height}{\includegraphics{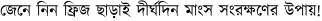}} (Explore how to preserve meat without a refrigerator!) \\

  Ambiguous & Imprecise or unclear titles that pique interest. &  \raisebox{-.26\height}{\includegraphics{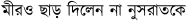}} (Not even Mir spared Nusrat) \\

  Exaggerate & Titles overstating what is written on the landing page. &  \raisebox{-.26\height}{\includegraphics{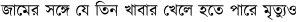}!} (Three foods when combined with blackberries, could kill you!)\\

  Graphic & Salacious, unsettling, or implausible subject matter. &  \raisebox{-.26\height}{\includegraphics{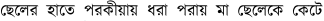}} \raisebox{-.26\height}{\includegraphics{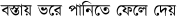}} (After he finds her cheating, the mother cuts her son into bits and stuffs him into a bag before tossing it into the water.)\\

  Formatting & Excessive use of punctuation or other symbols. &  \raisebox{-.26\height}{\includegraphics{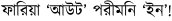}} (Faria 'out' Porimoni 'in'!)\\

  Bait \& Switch & Overpromising titles with under-delivering content; requires additional clicks. & \raisebox{-.26\height}{\includegraphics{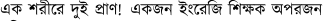}} \raisebox{-.2\height}{\includegraphics{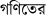}} (One body two souls! One is an English teacher, whereas another is of Mathematics.) \\
  
  \hline
\end{tabular}
\medskip
\caption{Clickbait news titles and their categories.}
\label{table:Clickbait categories}
\end{table*}

There are an estimated 11.4 million internet users in Bangladesh\footnote{\url{https://www.cia.gov/the-world-factbook/countries/bangladesh/}} who receive their daily news mostly from online news sites. However, no research has been conducted on tackling the increasing number of clickbait titles on these sites and other news websites. For English articles, \citet{potthast2018clickbait} built the first large-scale annotated clickbait corpus (Webis Clickbait Corpus 2017) containing 338,517 articles. In the Bangla language, the lack of an annotated clickbait-rich dataset is hindering the progress of Bangla Clickbait Detection. We construct the first Bangla Clickbait Corpus, which contains an article's title, content, and other metadata collected from various clickbait-rich websites upon which future researchers can build an effective Bangla clickbait detection model. The effectiveness of Semi-Supervised Generative Adversarial networks \citep[SS-GANs;][]{https://doi.org/10.48550/arxiv.1606.03498} have been shown for text classification tasks in \citet{croce-etal-2020-gan}. From our experiments, it is evident that fine-tuning a Bangla ELECTRA model in this setup improves clickbait detection performance outperforming all other model types.

The main contributions of this paper can be summarized as follows:
\begin{itemize}
\itemsep0em 
  \item We create an annotated dataset of 15,056 articles and an unannotated dataset of 65,406 Bangla articles rich with clickbait titles. The dataset contains the title, body, domain, article category, publication date, and English translation of title and content. We have released both of these datasets and the corresponding code to reproduce our experiments.
  \item We develop the first Bangla Clickbait Detection model for Bangla textual data by thoroughly experimenting with different statistical machine learning algorithms, deep neural networks using state-of-the-art embeddings, and Transformer networks \citep{vaswani2017attention} to discover the best approach for detecting clickbait. Section \ref{sec:result} analyzes the quantitative comparisons among all these different models. 
  \item We train a Bangla Transformer model in a Semi-Supervised Generative Adversarial setup and show that it improves upon existing models trained in a supervised manner. 
\end{itemize}

\section{Related Work}\label{sec:related_work}

The origin of clickbait is rooted in tabloids which have been in journalism since the 1980's \citep{bird2008tabloidization}. Generally, clickbait detection features can be obtained from 3 different origins: clickbait teaser phrase or post text, the attached article that the post text wants the user to click, and metadata for both \citep{potthast2018clickbait}. Apart from the post text, which is used by most to identify clickbait, the works of \citet{potthast2016clickbait} and \citet{biyani20168} also considered the linked article, metadata and used handcrafted features, TF-IDF similarity between headline and article content and Gradient Boosted Decision Trees (GBDT). \citet{potthast2018clickbait} suggested that clickbait detection should be a regression problem instead of a binary classification challenge, as the latter provides a way to measure how much clickbait is in the teaser message. They initiated the Webis clickbait challenge 2017, which boosted research activity in clickbait detection giving rise to highly effective and flexible deep learning techniques. For clickbait challenge 2017, \citet{zhou2017clickbait} first used self-attentive RNN \citep{elman1990finding} to select the important words in the title and created a BiGRU \citep{https://doi.org/10.48550/arxiv.1409.1259} network to encode the contextual information. \citet{thomas2017clickbait}, on the other hand, incorporated article content into an LSTM model \citep{6795963} for the clickbait challenge. \citet{rony2017diving} used continuous skip-gram model \citep{mikolov2013distributed} to generate the word embedding of clickbait titles. However, \citet{indurthi2020predicting} first investigated the application of transformer regression models in clickbait detection and achieved the first position in the clickbait challenge. Besides, \citet{hossain2020banfakenews} created the first Bengali newspaper dataset for Bengali fake news detection containing an annotated dataset of $\approx50K$ Bangla news. To the best of our knowledge, the first attempt to detect clickbait in Bangla was made by \citet{9528095}. They created a dataset on video-sharing platforms containing Bangla and English video links and used numerical features to detect clickbait links. However, no research has been conducted to 
 tackle clickbaits in written news mediums using the textual features of the article. We present the first clickbait detection dataset in Bangla and also provide a comprehensive comparison of various models to detect them.

\section{First Bangla Dataset for Detecting Bangla Clickbait News Articles}\label{sec:dataset}
\subsection{Data Collection}\label{sec:data_collection}

We first compile a list of websites that publish Bangla news articles. Although \citet{potthast2018crowdsourcing,potthast2016clickbait} used metrics like the number of retweets to select the most influential websites, such metric providing services like Alexa ranking\footnote{\url{https://www.alexa.com/}} is unavailable for most prominent Bangla Websites. Instead, we first create a preliminary list of Bangla news article sites from where we choose a website for scraping if the homepage seems to contain more clickbait than non-clickbait titles after a cursory glance by the annotators. We also select some famous Bengali online news publishers such as Kaler Kantha\footnote{\url{https://www.kalerkantho.com/}}, SomoyTV\footnote{\url{https://www.somoynews.tv/}}, and RTV news\footnote{\url{https://www.rtvonline.com/}} for scraping to facilitate future investigation into clickbait practices in popular Bangla news mediums. Before scraping, we check whether the publishers we select have terms and conditions against scraping or using their content for educational or research purposes to avoid copyright infringement. Utilizing the Python Selenium module, we have scraped data from the first week of February 2019 to the last week of February 2022. 

Although \citet{hossain2020banfakenews} published the first dataset of Bangla Fake news, we find it necessary to create a separate dataset for clickbait in Bangla as a news title can be a clickbait without necessarily being fake news \citep{10.1007/978-3-030-16145-3_5} \footnote{More details can be found in section \ref{sec:appendix_dif_clickbait_fake}}. To enrich our dataset size, one thousand titles labeled 'clickbait' from Bangla Fake News Dataset \citep{hossain2020banfakenews} are added to our own dataset after their labels are revised again by annotators.

\begin{table*}
\begin{minipage}{\columnwidth}
    \centering
     \resizebox{1\columnwidth}{!}{%
        \begin{tabular}{@{}lll@{}}
        \hline
        \textbf{Information} & \textbf{Value} \\
        \hline
        Crawling Period & Feb 2019 - Feb 2022\\
        Total Clickbait &  5239\\
        Total Non-clickbait &  9817\\
        Total Unlabelled &  65406\\
        \hline
        \multirow{2}{*}{\textbf{Title Analysis}} & \multirow{2}{*}{\textbf{Clickbait}} & \multirow{2}{*}{\textbf{Non-clickbait}}\\
                          &                            &\\ \hline
        Average number of characters    & 52.845                & 49.097\\
        Average number of words         & 8.983                & 7.8356\\
        Average word length & 4.99                & 5.4\\
        Average Punctuation  &  1.003      &  0.805                \\
        \hline
        \end{tabular}
     }
     \caption{Summary statistics of our dataset.}
     \label{table:summary statistics}
\end{minipage}
\hfill 
\begin{minipage}{\columnwidth}
    \centering
    \resizebox{0.9\columnwidth}{!}{%
        \begin{tabular}{@{}ll@{}}
        \hline
        \textbf{Column} & \textbf{Value}\\
        \hline
        Domain & \url{https://www.rtvonline.com/}\\
        Date &  2021-05-25\\
        Title &  
        \raisebox{-.3\height}{\includegraphics{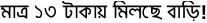}}
        \\
        Content & \raisebox{-.8\height}{\includegraphics{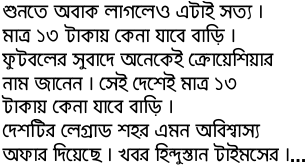}}
        \\
        Label & 1 (Clickbait) \\
        Translated Title & It's only Rs. 13!\\
        Translated Content & \raisebox{-.6\height}{\shortstack{That's the truth, though it sounds surprising.\\
        Only 13 rupees can be bought at home...}}\\
        Category &  Science \& Technology\\
        \hline
        \end{tabular}
       
    }
     \caption{Sample Data}
     \label{data instance and statistics table}
  \end{minipage}
\end{table*}

\subsection{Annotation Process}\label{sec:annot_process}
The dataset is annotated by three annotators with an MA in Bangla Linguistics. At first, they study the annotations of popular English clickbait datasets \citep{potthast2018crowdsourcing, Agarwal2016Clickbait,potthast2016clickbait}. Investigating
English titles help the annotators understand how titles induce curiosity in practice, which they can then use to annotate Bangla titles. As questions naturally entice interest, a new clickbait category named 'question' is added to the clickbait categories in Table \ref{table:Clickbait categories}. No publisher or source of the article is available to the annotators to avoid any induced publisher-based biases as reported by \citet{potthast2018clickbait} to be the case for several clickbait datasets \citep{rony2017diving,Chakraborty2016Stop,Agarwal2016Clickbait}. A majority vote among the annotators decides the final annotation. The annotators reach an inter-annotator agreement Fleiss kappa \citep{fleiss1971mns} of 0.62, which is substantial  \citet{10.2307/2529310} and enough for a good speculative conclusion regarding annotator agreement \citep{artstein-poesio-2008-survey}.

The annotators mark clickbait news as a numeric value of 1 and non-clickbait news as a numeric value of 0. Our labeled and unlabelled datasets contain eight categories - Economy, Education, Entertainment, Politics, International, Sports, National, and Science \& Technology of clickbait and non-clickbait titles. After removing all duplicates from labeled and unlabeled datasets, our dataset contains 15,056 unique news articles with 9,817 non-clickbait and 5,239 clickbait articles, and 65,406 unique unlabelled articles. The labeled and unlabeled datasets do not have any overlapping content or titles. The test set is further curated by removing titles that have similar titles in the training set through Levenshtein distance \citep{Levenshtein1965BinaryCC}. Table \ref{table:summary statistics} shows that clickbait titles have a slightly higher average number of words and punctuation than non-clickbait titles. The most frequent fifteen words in clickbait titles are - 
\par \includegraphics[scale=0.85]{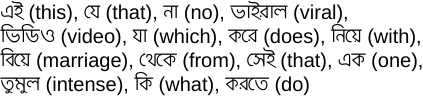}

It contains words like viral, video, and intense which usually induce readers to click. Each data instance contains the title and content of the article, publishing date, domain, news category, translated title, and translated content as shown in Table \ref{data instance and statistics table}. 

\section{Human Baseline}\label{sec:baseline}
Five human annotators who are undergraduate and regular newspaper readers are given 200 news article titles from the test set to annotate. They achieve an inter-annotator agreement of Fleiss' kappa \citep{fleiss1971mns} score of 0.374, which is fair according to \citet{10.2307/2529310}. Compared to our dataset annotators' score, this score is much lower. Our annotation process includes investigating English titles first to better form a coherent perception of clickbait titles. By majority voting among the five annotators, we select the final labels and achieve an F1 score of 76.82\%  and an accuracy of 77.01\% on the clickbait class, which serves as the human baseline for Bangla clickbait detection shown in  Table \ref{table: all model results}.
\section{Approach}\label{sec:approach}

\subsection{GAN-BanglaBERT}\label{ss-gan}
\begin{figure*}[!htb]
  \centering
      \centering
      \includegraphics[width=1\textwidth]{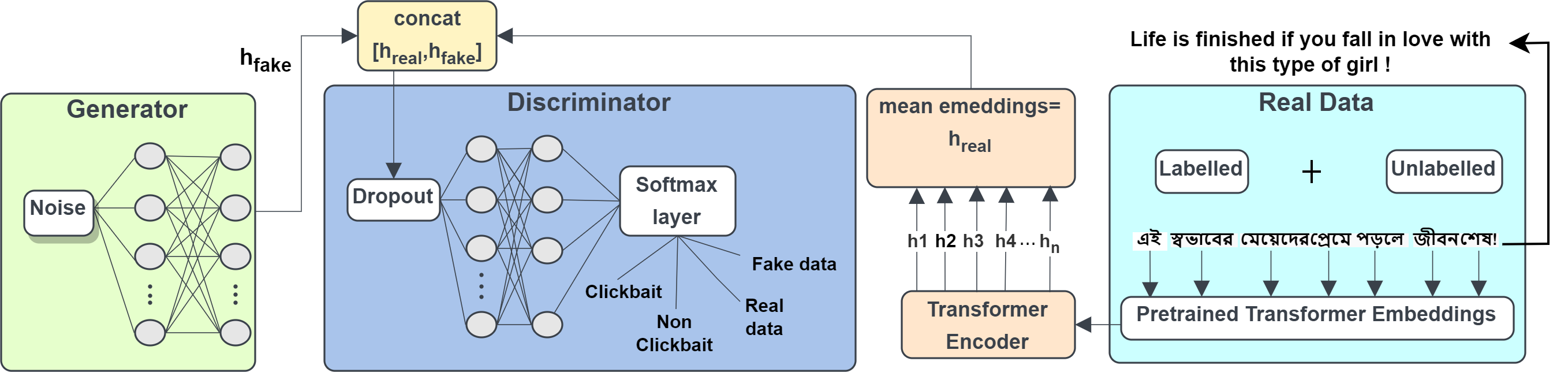}
      \caption{{\tt GAN-BanglaBERT} architecture. Generator $\mathcal{G}$ generates fake data given random noise, and Discriminator $\mathcal{D}$ takes both real and this fake data and outputs four labels: 0 for non-clickbait, 1 for clickbait, 2 for real and 3 for fake data.}
      \label{fig: Gan model}
\end{figure*}

In Generative Adversarial Network \citep{https://doi.org/10.48550/arxiv.1406.2661}, a generator $\mathcal{G}$ is trained to generate a data distribution similar to the real data to 'fool' the discriminator $\mathcal{D}$ and $\mathcal{D}$ is trained to differentiate between the two in an adversarial fashion. Semi-Supervised GANs \citep[SS-GANs;]{https://doi.org/10.48550/arxiv.1606.03498} train the discriminator $\mathcal{D}$ to predict the classification labels along with the additional task of predicting whether the data is real or fake. This training technique helps the model improve its inner representations by utilizing the unlabelled and generated data \citep{croce-etal-2020-gan}. Following researchers of \citet{croce-etal-2020-gan}, we finetune a BanglaBERT \citep{bhattacharjee2021BanglaBERT}, a state-of-the-art ELECTRA \citep{clark2020electra} model pre-trained on 35 GB of Bangla textual data from the web and call it 'GAN-BanglaBERT' throughout the paper. Figure \ref{fig: Gan model} shows the overall architecture of the GAN-BanglaBERT model. Generator $\mathcal{G}$ and discriminator $\mathcal{D}$ both are a 2-layered deep neural network(DNN). A 100-dimensional noise vector is drawn from a standard normal distribution $N\left(\mu=0, \sigma^2=1\right)$ following the initialization practice in GANs \citep{https://doi.org/10.48550/arxiv.1406.2661}. Generator $\mathcal{G}$ produces $h_{\text{fake }} \in \mathbb{R}^d$ vector from this noise vector where $d$ is the last layer size of the pre-trained Transformer network. Discriminator $\mathcal{D}$ takes in input the concatenation of both real and fake data's representation $[h_{\text{real }};h_{\text{fake}}]$. Detailed training loss calculation is provided in \citet{croce-etal-2020-gan}, which remains unchanged in our implementation. The average of the last hidden layer outputs of BanglaBERT is the transformer encoding $h_{\text{real }}$ for a real title.

\subsection{Comparison Methods}\label{compare-methods}
We compare the GAN-BanglaBERT model to the following models.
\begin{itemize}
\item[-] Statistical Models: For statistical methods, we employ a Logistic and Random Forest classifier on a combination of various features like TF-IDF (term frequency–inverse document frequency) of the word and character n-grams (n-gram range=3-5), Bangla pre-trained word embeddings, punctuation frequency, and normalized \textit{Pars-of-Speech} frequency according to \citet{hossain2020banfakenews}.

\item[-] \citet{zhou2017clickbait}: employ a BiGRU \citep{https://doi.org/10.48550/arxiv.1409.1259} network with a self-attentive network \citep{yang-etal-2016-hierarchical} on top of the BiGRU representations and achieve the first position at Clickbait Challenge 2017 \citep{potthast2018clickbait} with an F1 score of 0.683. 

\item[-] \citet{Agarwal2016Clickbait}:  employ a multi-channel CNN model with one convolutional layer similar to the model demonstrated by \citet{kim2014convolutional}. Pre-trained word embeddings are passed to multiple filters, and their concatenated representation is sent to a Max Pooling layer for the final representation. 

\item[-] \citet{https://doi.org/10.48550/arxiv.2104.05243}:
We translate all our article titles using a Bangla-to-English translator model Bangla-NMT \citep{https://doi.org/10.48550/arxiv.2009.09359} which outperformed Google Translate on SUPara-benchmark test set \citep{9038381}. The translated titles are passed into a state-of-the-art misinformation detection model UnifiedM2 \citep{https://doi.org/10.48550/arxiv.2104.05243} trained on fake, clickbait, rumor, and news-bias datasets in English. We investigate if translating the titles and using a state-of-the-art model trained in English suffices for clickbait detection or whether language-specific training is necessary.
\end{itemize}

\section{Experimental Setup}
\subsection{Pre Processing}\label{pre-processsing}
Normalizer module by \citet{hasan-etal-2020-low} and Bangla unicode normalizer by \citep{alam2021large} are used for Unicode and nukta normalization, removing HTML tags, URL links, etc. High punctuation usage is a common trait of clickbait titles. We preserve all syntactically correct punctuation in our titles and remove punctuation that appeared in the middle of words causing words to break and create out-of-vocabulary words for models.

\subsection{Experimental Settings}\label{model-config}
For all models, we use the article's title as input as the title mainly creates the curiosity gap that is the principal characteristic of a clickbait title \citep{potthast2016clickbait}. We use Bangla Fasttext \citep{bojanowski-etal-2017-enriching} and Bangla Word2Vec embedding pre-trained on Bangla Wikipedia Dump Dataset with coverage of 65.16\% and 60.91\% respectively, on the total vocabulary size of article titles as embedding inputs. We extract the \textit{Parts of Speech} (POS) tags using BNLP toolkit \citep{sarker2021bnlp}. We derive a Bangla punctuation list from \citet{alam2021large}. We experiment with both BiGRU and BiLSTM models for \citep{zhou2017clickbait} model and show the better performing one in section \ref{sec:result}. The above models are trained for 40 epochs with Adam optimizer \citep{kingma2017adam} and learning rate = 2\textit{e}-5, which is changed dynamically according to 1cycle learning rate scheduler \citep{smith2018superconvergence}. The GAN-BanglaBERT and BanglaBERT models are trained for 20 epochs with AdamW optimizer \citep{loshchilov2019decoupled}, and the learning rate is slowly increased from zero to 1\textit{e}-5 within a warmup period. For GAN-BanglaBERT, the learning rate for the generator and discriminator model is kept the same. For all models, we pad or truncate titles to lengths of 64. The labeled dataset is split into 70:10:20 fashion for training, validation, and test splits using stratified sampling. All models are trained with batch size=64, and the best model based on the validation result is used to evaluate the final test set. Each experiment is repeated five times, and the average result on the held-out test set is used for the final result of all the models.

\section{Results and Analysis}\label{sec:result}

\begin{table*}[htb!]
\centering
\begin{tabular}{@{}lrrrrr@{}}
\hline
\textbf{Model}            & \textbf{F1 Score} & \textbf{Precision} & \textbf{Recall} & \textbf{Accuracy} \\ \hline
\citet{zhou-etal-2016-attention} ({\tt Fasttext})           & 39.37                        & 39.88                         & 38.87                      & 57.87                     \\
\citet{Agarwal2016Clickbait} ({\tt Fasttext})     & 35.15                         & 40.05                         & 31.32                      & 59.33                 \\
Logistic Regression ({\tt character 3,4,5 gram})                       & 66.28                         & 75.36                          & 59.15                       & 78.82                     \\
Random Forest ({\tt character 3 gram})                       & 67.01                         & 61.06                          & 74.25                       & 74.27                \\
\citet{https://doi.org/10.48550/arxiv.2104.05243}                   & 11.02                         & 39                         & 6.4                         & 63.53                 \\
BanglaBERT                      & 71.72                         & 80.42                          & 64.71                       & 82.04                 \\
GAN BanglaBERT                  & \textbf{75.13}                           & \textbf{75.45}                         & \textbf{74.81}                       & \textbf{82.57}                         \\
Human Baseline                  & 76.81                           & 77.6                         & 76.04                       & 77.01                        \\

 \hline
\end{tabular}
\caption{Performance comparison of GAN-BanglaBERT and all other models on the test set. F1 score, precision, and recall are for the clickbait class. For \citet{zhou2017clickbait} model, a better performing BiLSTM-attn model result is shown. GAN-BanglaBERT outperforms all other models and the performance difference is statistically significant ($p < 0.01$) according to McNemar’s test \citep{10.1162/089976698300017197}}
\label{table: all model results}
\end{table*}
Table \ref{table: all model results} illustrates the performance of all models on our test set. For each type of model, only the best-performing feature's result is shown. GAN-BanglaBERT outperforms all other models regarding F1 score, precision, metric, and recall. It achieves a 75.13\% F1 score on the clickbait class, which is 3.41\% greater than the supervised BanglaBERT model. The performance is close to the human upper bound of  76.8\% F1 score. The human baseline score shows that separating clickbait and non-clickbait titles is a difficult task even for humans, and clickbait may not be perceptible to all humans \citep{potthast2018crowdsourcing}. 

\begin{figure}[!htb]
\includegraphics[width=\columnwidth]{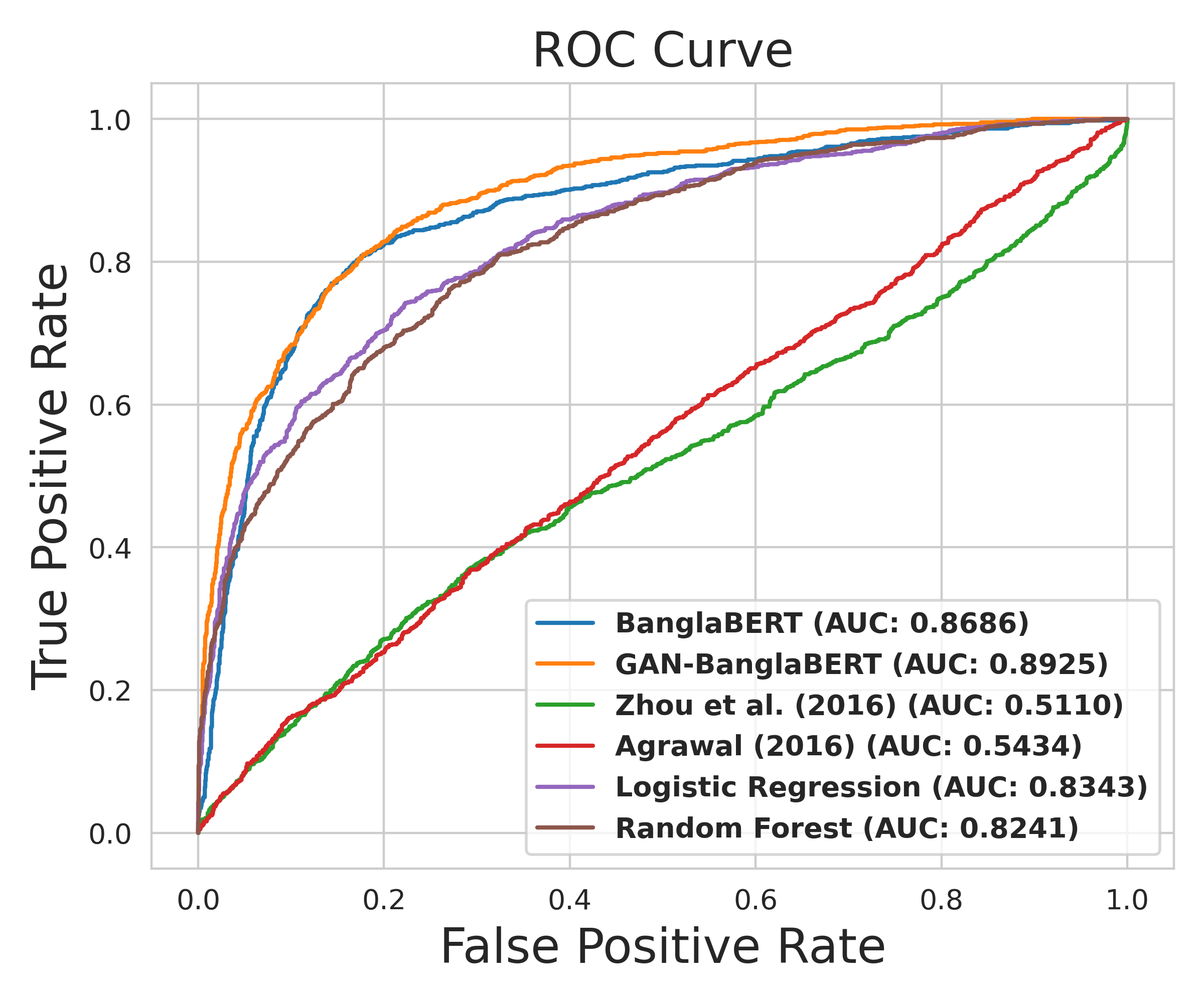}
\caption{ROC curve for all models where GAN-BanglaBERT achieves the highest Area Under ROC Curve (AUC) score.}
\label{fig:roc-curve}
\end{figure}

Figure \ref{fig:roc-curve} shows the ROC curve (receiver operating characteristic curve) for all models where the GAN-BanglaBERT model achieves the highest AUC (area under ROC curve) score of 0.8925, which is higher than the BanglaBERT. The high AUC score of GAN-BanglaBERT suggests that it can distinguish between clickbait and non-clickbait titles more accurately than other models. 

\begin{figure*}[hbt!]
     \centering
     \begin{subfigure}[b]{0.4\textwidth}
         \centering
         \includegraphics[width=\textwidth]{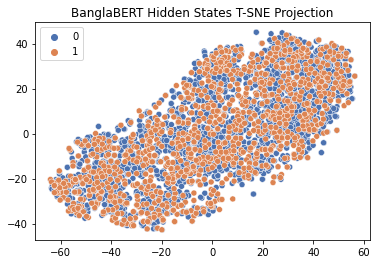}
         \caption{BanglaBERT t-SNE}
         \label{fig:transformer_tsne}
     \end{subfigure}
     \begin{subfigure}[b]{0.4\textwidth}
         \centering
         \includegraphics[width=\textwidth]{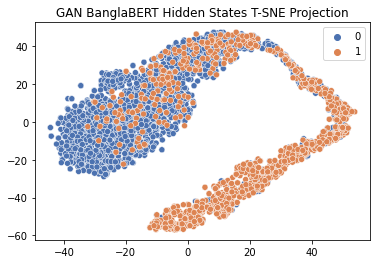}
         \caption{GAN-BanglaBERT t-SNE}
         \label{fig:gan_tsne}
     \end{subfigure}
     \caption{Visualization of last layer hidden representations using t-SNE for BanglaBERT (\ref{fig:transformer_tsne}) and for GAN-BanglaBERT (\ref{fig:gan_tsne}). 0 represents Non-Clickbait and 1 represents Clickbait in both figures.}
     \label{fig:tsne_gan_banglabert}
\end{figure*}

To investigate whether training in a semi-supervised approach improves BanglaBERT's inner representations as stated by \citep{croce-etal-2020-gan}, we plot the average of the last layer hidden representations of GAN-BanglaBERT and BanglaBERT using a t-SNE projection \citep{JMLR:v9:vandermaaten08a} in Figure \ref{fig:gan_tsne}. GAN-BanglaBERT better separates the clickbait class from the non-clickbait than the BanglaBERT model, proving that training a BERT model in a semi-supervised adversarial manner can improve the learned representations of the model and thus improve performance. 

\begin{table*}[!htb]
\centering
\small
\begin{tabular}{llll}
\hline
\textbf{Category} & \multicolumn{2}{l}{\textbf{Attention Weighted Words}} & \textbf{Important Words}\\ \hline
\multirow{4}{*}{Ambiguous} & BanglaBERT & \raisebox{-.3\height}{\includegraphics{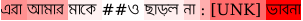}}          & \multirow{4}{*}{\raisebox{-.3\height}{\includegraphics{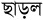}}, \raisebox{-.3\height}{\includegraphics{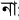}} (forsake, not)} \\
& GAN-BanglaBERT & \raisebox{-.3\height}{\includegraphics{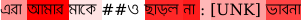}} &                     \\ 
& Title & \raisebox{-.3\height}{\includegraphics{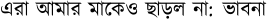}} &                     \\
& Translation & They did not even forsake my mother &                     \\
\hline
\end{tabular}
\medskip
\caption{Comparison between GAN-BanglaBERT and BanglaBERT on ambiguous type clickbait title prediction. Each word is highlighted according to the attention weight given by the model.}
\label{fig:x Error Analysis}
\end{table*}

For creating the unlabelled dataset, we choose clickbait-dense websites from the web to ensure a higher abundance of clickbait titles. To investigate whether this helps performance, we create another unlabelled dataset of the same size from Daily Prothom Alo archive\footnote{https://github.com/zabir-nabil/bangla-news-rnn}, which has a substantially lower clickbait ratio. Our model achieves 72.38\% F1 score on this second unlabelled set compared to 75.13\% F1 score on the original unlabelled set, proving that a higher clickbait ratio in the unlabelled set improves performance on the Clickbait class. 

Table \ref{fig:x Error Analysis} shows a prominent clickbait category - 'ambiguous' where GAN-BanglaBERT performs better than other models. 'They did not even forsake my mother! - Bhabna' is a quotation that implies something ostentatious happened with the mother, although expressed very vaguely. {\raisebox{-.3\height}{\includegraphics{pdfs/att_2.pdf}}} \raisebox{-.3\height}{\includegraphics{pdfs/att_3.pdf}} (not, forsake) words create this ambiguity which GAN-BanglaBERT correctly gives more attention to, but BanglaBERT fails to do so. The high AUC score and better separation in encoding shown in Figure \ref{fig:tsne_gan_banglabert} enables GAN-BanglaBERT to perform better in these harder-to-detect cases.

Table \ref{table: all model results} shows that \citet{https://doi.org/10.48550/arxiv.2104.05243} model on translated titles performs very poorly compared to other models. Machine translation produces more synthetic text, which diminishes the lexical and syntactical style and richness of the source language \citep{vanmassenhove-etal-2021-machine}. For example,
\raisebox{-.6\height}{\includegraphics{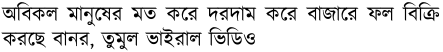}}

is translated to 'Monkeys selling fruit in the market at the expense of the real man, viral video.' Although this translation is factually correct, it loses the source language's exaggerated tone, leading to misclassification.

Logistic regression and Random Forest model on character TF-IDF features heavily outperform neural network models like BiLSTM with attention network and CNN \citep{zhou2017clickbait,Agarwal2016Clickbait}. These models can effectively identify certain keywords that are very significant in classifying clickbait titles. For instance, a top character feature returned by logistic regression is \raisebox{-.1\height}{\includegraphics{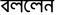}} (told), which is a common keyword found in many clickbait titles, e.g., \raisebox{-.1\height}{\includegraphics{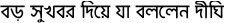}} (What Dighi said about the great news). The poor performance of neural network models can be attributed to Bangla pre-trained Fasttext and Word2Vec embeddings, which are trained on the Bangla Wikipedia dump and are significantly smaller in size than English. Training these embeddings on training data and then initializing the neural models with these embeddings may improve performance.

All models perform poorly on Bait \& Switch type titles as mentioned in Table \ref{table:Clickbait categories} where titles where the main content under-delivers the title's statements. As these types of clickbait require reading the content to predict correctly, all models underperform as they are trained on only the article's title. Effectively combining content features with titles to classify these types of clickbait titles is a future research endeavor for us.

\section{Conclusion}
We present the first clickbait detection dataset containing 15,056 labeled new articles and 65,406 unlabelled articles containing article title, content and metadata to enable researchers to use this dataset to build state-of-the-art clickbait detection models. By conducting a comprehensive study on various architectures, we provide a strong baseline for detecting clickbait in Bangla articles. We show that training a pre-trained Transformer model in a semi-supervised approach by incorporating unlabeled data improves performance and inner representation. As simple statistical models perform strongly on clickbait titles, we aim to investigate how these features can be combined with word embeddings to pass into neural networks. We also plan to investigate how features from article content can be utilized to detect clickbait. We wish to publicly release the dataset and code to further progress into Bangla clickbait detection.

\section*{Acknowledgments}

We thank our data annotators and the volunteers from the Department of CSE, BRAC University, who participated in the human baseline experiment. We also thank the CSE Dept. of BRAC University, Bangladesh, for their continued support and direction. We show our sincerest gratitude to the anonymous reviewers and the pre-submission mentors for their valuable suggestions which helped improve the research work.

\bibliographystyle{acl_natbib}
\bibliography{custom}

\appendix
\section{Appendix}
\label{sec:appendix}
\subsection{Data sources}
\label{appendix:data_sources}
We choose a site for scraping if the homepage seems to contain more
clickbait than non-clickbait titles after a cursory
glance by the annotators.  We also select some famous Bengali online news publishers such as Kaler
Kantha4, SomoyTV5, and RTV news6
for scraping to facilitate future investigation into clickbait
practices in popular Bangla news mediums. Table \ref{table:appendix_data_sources} contains the data sources for our dataset's labeled and unlabelled portions.

\begin{table}[htb!]
\resizebox{\columnwidth}{!}{%
\begin{tabular}{llr}
\hline
\multicolumn{3}{c}{Labelled}                             \\ \hline
\multirow{2}{*}{Domain} & \multicolumn{2}{c}{News Count} \\
                        & clickbait    & non-clickbait   \\ \hline
twentyfourbd            & 1727         & 1062            \\
topdhaka                & 1004         & 920             \\
rtvonline               & 1003         & 86              \\
BanFakeNews             & 623          & 415             \\
kureghornews            & 634          & 375             \\
newzcitizen             & 633          & 351             \\
nbtimes24               & 750          & 228             \\
citynewszet             & 503          & 451             \\
authoritynewz           & 561          & 385             \\
thebasenewz             & 537          & 268             \\
newzauthority           & 308          & 281             \\
newsholder21            & 361          & 117             \\
techzoom                & 369          & 60              \\
channeldhaka            & 291          & 38              \\
kalerkantho             & 285          & 37              \\
somoynews               & 194          & 38              \\
beanibazarview24        & 109          & 52              \\ \hline
\end{tabular}%
}
\resizebox{\columnwidth}{!}{%
\begin{tabular}{ccc}
\multirow{10}{*}{Unlabelled} & \multirow{2}{*}{Domain} & \multirow{2}{*}{News Count} \\
                             &                         &                             \\ \cline{2-3} 
                             & mtnews24                & 16194                       \\
                             & dakpeon24               & 1567                        \\
                             & newsfastcreator         & 8836                        \\
                             & propernewsbd            & 1830                        \\
                             & thecityvpn              & 14455                       \\
                             & usbanglanews            & 16099                       \\
                             & glamourbd               & 6197                        \\
                             & jagonews24              & 228                         \\ \cline{2-3} 
\end{tabular}%
}
\caption{Data sources of Bangla Clickbait Datset}
\label{table:appendix_data_sources}
\end{table}
\subsection{Detailed Results}
\label{appendix:result_section}

For statistical models, we experimented with Random Forest and Logistic Regression networks. We passed various types of lexical, syntactical, and embedding features to these networks to investigate which performs best. For neural network models, we employ architectures from two previous research works \citealp[]{zhou-etal-2016-attention,Agarwal2016Clickbait}. For Transformer networks, we train commonly available Bangla pre-trained transformer models in both classic and semi-supervised GAN manner. Table \ref{table:appendix_all_experiments} contains the results of these experiments.

\begin{table}
\resizebox{\columnwidth}{!}{%
\begin{tabular}{>{\centering\hspace{0pt}}m{0.425\linewidth}>{\centering\hspace{0pt}}m{0.285\linewidth}>{\centering\arraybackslash\hspace{0pt}}m{0.221\linewidth}}
\multicolumn{3}{>{\centering\arraybackslash\hspace{0pt}}m{0.93\linewidth}}{\textbf{Statistical Classifiers }} \\ 
\hline
\textbf{Traditional Linguistic Features} & \textbf{Logistic Regression} & \textbf{Random Forest} \\ 
\hline
Unigram (U) & 57.39 & 56.11 \\
Bigram (B) & 29.7 & 53.34 \\
U+B+T & 57.68 & 55.94 \\
C-3 gram & 64.81 & \textbf{67.01} \\
C-4 gram & 65.59 & 62.48 \\
C-5 gram & 65.13 & 58.58 \\
C3+C4+C5 & \textbf{66.28} & 65.36 \\
All Lexical(L = U+B+T+C3-C5) & 64.29 & 65.6 \\
Parts of Speech(POS) & 33.14 & 40.37 \\
L+POS & 62.23 & 65.97 \\
Embedding Word2Vec (E W) & 53.11 & 51.35 \\
Embedding Fasttext (E F) & 50.19 & 49.4 \\
L+POS+E W & 64.2 & 65.04 \\
L+POS+E F & 64.43 & 65.05 \\
Punctuation (P) & 5.88 & 52.06 \\
\textcolor[rgb]{0.075,0.075,0.075}{L+POS+E W+P} & \textcolor[rgb]{0.075,0.075,0.075}{63.34} & 64.7 \\
L+POS+E N+P & 64.34 & 64.91 \\
All features & 64.73 & 63.26 \\
\hline
\end{tabular}
}
\resizebox{\columnwidth}{!}{%
\begin{tabular}{>{\centering\hspace{0pt}}m{0.598\linewidth}>{\centering\hspace{0pt}}m{0.15\linewidth}>{\centering\arraybackslash\hspace{0pt}}m{0.163\linewidth}} 
\hline
\textbf{Transformer Networks} & \textbf{Classic} & \textbf{SS-GAN} \\ 
\hline
BERT base multilingual cased & 62.37 & 70.21 \\
Bangla BERT Base & 68.13 & 68.54 \\
Indic-BN-BERT & 72.21 & 73.36 \\
Indic-BN-RoBERTa & 67.76 & 70.52 \\
DistilBERT base multilingual cased & 69.61 & 70.38 \\
Indic-BN-DistilBERT & 71.32 & 72.35 \\
Bangla-Electra & 66.79 & 67.77 \\
Indic-BN-XLM-RoBERTa & 71.82 & 70.75 \\
CSENLP-BanglaBert & \textbf{71.72} & \textbf{75.13} \\
CSENLP-BanglaBert\_Large & 71.66 & 72.07 \\
\hline
\end{tabular}
}
\resizebox{\columnwidth}{!}{%
\begin{tabular}{>{\centering\hspace{0pt}}m{0.591\linewidth}>{\centering\arraybackslash\hspace{0pt}}m{0.200\linewidth}} 
\hline
\multicolumn{2}{>{\centering\arraybackslash\hspace{0pt}}m{0.791\linewidth}}{Neural Networks} \\ 
\hline
CNN \citep{Agarwal2016Clickbait} & 35.15 \\
Bi-LSTM \citep{zhou-etal-2016-attention}  & 39.37 \\
\hline
\end{tabular}
}
\caption{Detailed result of all experiments conducted on BanglaBait dataset}
\label{table:appendix_all_experiments}
\end{table}

\subsection{Difference between Clickbait and Fake news}\label{sec:appendix_dif_clickbait_fake}
Although \citet{hossain2020banfakenews} published the first dataset of Bangla Fake news, we don't focus on the misinformation, fabricated or fake content within the articles, or their authenticity to detect clickbait in this dataset. The following two examples explain the difference between fake and clickbait titles in detail- \\
Example 1: Buying land on the Moon is the current craze. Explore how you can do that too!\\
\includegraphics{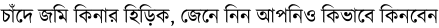}\\
Example 2: 'Hawa' got nominated for the Oscars\\
\includegraphics{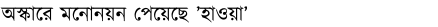}\\
Example 1 presents an accurate title (verified by renowned news publishers such as the Kalerkantho and the Somoynews) in a clickbait-style by using hyperbolic words like 'craze' and alluring phrases like 'Explore how you can do that too.' It proves a clickbait article does not have to be fake to be clickbait. Example 2, on the other hand, is fake news verified from the official Facebook page of the movie 'Hawa', however, the title style is not exactly luring readers to click, proving that an article can be fake without being clickbait. In short, clickbait headlines do not necessarily have to be fake news; they may contain genuine information but in an exaggerated fashion \citep{10.1007/978-3-030-16145-3_5}. \citet{biyani20168} includes factually wrong articles in the 'wrong' category of clickbait articles.

\end{document}